\documentclass{article}
\usepackage{spconf,amsmath,graphicx}

\usepackage{lipsum}
\usepackage{graphicx}
\usepackage{bm,amssymb,mathrsfs,amsfonts}
\usepackage{caption}
\usepackage{scalefnt}
\usepackage{multirow}
\usepackage{enumitem}
\usepackage{type1cm}
\usepackage{cite}
\usepackage{algorithm}
\usepackage{algorithmic}
\usepackage{here}

\newcommand{\argmax}{\mathop{\rm arg~max}\limits}

\title{Improving OOV Detection and Resolution\\with External Language Models in Acoustic-to-Word ASR}
\name{Hirofumi Inaguma$^1$, Masato Mimura$^1$, Shinsuke Sakai$^1$, Tatsuya Kawahara$^1$}
\address{$^1$Graduate School of Informatics, Kyoto University, Japan}
\begin{document}
\ninept
\maketitle
\begin{abstract}
Acoustic-to-word (A2W) end-to-end automatic speech recognition (ASR) systems have attracted attention because of an extremely simplified architecture and fast decoding.
To alleviate data sparseness issues due to infrequent words, the combination with an acoustic-to-character (A2C) model is investigated.
Moreover, the A2C model can be used to recover out-of-vocabulary (OOV) words that are not covered by the A2W model, but this requires accurate detection of OOV words.
A2W models learn contexts with both acoustic and transcripts; therefore they tend to falsely recognize OOV words as words in the vocabulary.
In this paper, we tackle this problem by using external language models (LM), which are trained only with transcriptions and have better linguistic information to detect OOV words.
The A2C model is used to resolve these OOV words.
Experimental evaluations show that external LMs have the effects of not only reducing errors but also increasing the number of detected OOV words, and the proposed method significantly improves performances in English conversational and Japanese lecture corpora, especially for out-of-domain scenario.
We also investigate the impact of the vocabulary size of A2W models and the data size for training LMs.
Moreover, our approach can reduce the vocabulary size several times with marginal performance degradation.
\end{abstract}


\section{Introduction}\label{sec:intro}
The conventional HMM-based hybrid automatic speech recognition (ASR) systems are modularized into several components such as acoustic, subword, lexicon, pronunciation, and language model.
While they are shown to achieve a human-level recognition performance \cite{saon2017english, stolcke2017comparing}, they are separately optimized based on different objectives and training data, so the overall system is sub-optimal and needs a complicated decoding process to combine them.

In contrast, end-to-end systems, which optimize the direct mapping from acoustic features to transcriptions, have shown promising results using three end-to-end ASR models: connectionist temporal classification (CTC) \cite{ctc_2006,blstmctc_icassp2013, eesen, deepspeech2}, attention-based encoder-decoder models \cite{attention_nips2015, las, attention_icassp2016}, and RNN-transducer models \cite{seq2seq_comparison_is17, baidu_seq2seq_asru2017}.
In this paper, we focus on attention-based models because they show the most promising results among them \cite{seq2seq_comparison_is17, baidu_seq2seq_asru2017, ueno_icassp2018}.
There are some choices about output units, and a majority of the previous end-to-end systems is based on subword units such as characters \cite{las} and word-piece units \cite{google_sota_asr}.
Recently, acoustic-to-word (A2W) models have received much attention because of their extremely simplified architecture and fast decoding process \cite{fast_accurate,direct_a2w_is17,direct_a2w_icassp2018,nsr,a2w_without_oov_asru2017,lu2016training,ueno_icassp2018, advancing_ctc_icassp2018}.
When considering downstream processes of ASR such as machine translation, dialogue systems, and spoken term detection, word-level information is required, so it is a natural choice to adopt a word as the output unit.
However, A2W models need a considerable amount of training data to match the performance of the state-of-the-art subword-based models \cite{nsr}.
Therefore, they encounter problems of data sparseness and over-fitting due to infrequent words \cite{fast_accurate}.
Moreover, A2W models cannot recognize out-of-vocabulary (OOV) words.

To tackle these problems, some works explore model initialization \cite{direct_a2w_is17} and multi-task learning (MTL) with low-level auxiliary tasks \cite{a2w_without_oov_asru2017, ueno_icassp2018, advancing_ctc_icassp2018}.
In the MTL approach, we jointly optimize an A2W model with an acoustic-to-character (A2C) model by sharing encoder parameters \cite{a2w_without_oov_asru2017, ueno_icassp2018, advancing_ctc_icassp2018}.
OOV words are resolved by referring to the corresponding partial hypothesis provided by the A2C model \cite{a2w_without_oov_asru2017, ueno_icassp2018}.
It also has generalization effects by accelerating better parameter representations and leads to fast and stable convergence.

However, even with these regularization techniques, they are more likely to incorrectly recognize OOV words as other words in the predefined vocabulary.
This is because A2W models learn contexts with both acoustic and word sequence, and attach too much importance to acoustic level.
When recognizing speech corresponding to OOV words (these are often infrequent words), the A2W model assign high probabilities to other in-vocabulary words which have similar pronunciations.
In this case, these words cannot be recovered by the A2C model.

External language models (LM) trained with a large text have better linguistic information and further improve the ASR performance.
They are integrated to the attention-based models in the beam search decoding, which are referred to as \textit{shallow fusion} \cite{kannan2017analysis}.
Since external LMs are trained with a large text, they have better ability to detect OOV words accurately.

In this paper, we improve the OOV resolution for the attention-based A2W model by integrating the external LM.
We combine \textit{shallow fusion} with external LMs and the OOV resolution method by referring to the hypothesis of the A2C model.
We show that external LMs not only reduce errors of the A2W models but also increase the number of OOV words in the hypothesis, which means that more OOV words are detected with external LMs.
Then, recovering these OOV slots by the A2C model boosts the performance.
In addition, we found that the effect of external LMs are enhanced with MTL without recovering OOV words.

We conduct experimental evaluations with English conversational and Japanese lecture corpora, and show the proposed method significantly improves the performances, especially for out-of-domain test sets.
Moreover, we investigate the effects of the vocabulary size of the A2W models and the data size for training external LMs.
By restricting the vocabulary to the frequent words and recovering many OOV words from the A2C model, the vocabulary size can be reduced several times from the best model with marginal performance degradation.

The remaining part of this paper is structured as follows.
In Section \ref{sec:related_work}, we describe previous research on OOV problems of A2W models.
In Section \ref{sec:model}, we describe attention-based A2W models and the proposed method.
Experimental evaluations are presented in Section \ref{sec:experiment}.
We conclude this paper in Section \ref{sec:conclusions}.

\section{Acoustic-to-word end-to-end speech recognition without OOV}\label{sec:related_work}
Acoustic-to-word (A2W) end-to-end systems are attractive because they can directly optimize mapping from acoustic to word sequences with a single architecture, and realize fast decoding.
Although A2W models have these advantages, they suffer from the data sparseness problems due to rare words and therefore require sufficient training data.
Moreover, A2W models cannot recognize any OOV words.
A practical solution is to adopt word-pieces as output units \cite{google_sota_asr}.
There are some solutions to make A2W models open-vocabulary.
The first one is to decompose infrequent words to a sequence of subwords and encapsulate them into a single dictionary \cite{advancing_ctc_icassp2018, direct_a2w_icassp2018}.
Another solution is to recover OOV words by referring to the hypothesis of the A2C model, which is realized by joint training\cite{a2w_without_oov_asru2017, ueno_icassp2018}.
Time alignments of the A2W and A2C models are synchronous because they share encoder parameters.
This is similar to human perception, where frequent words are memorized but rare words such as named entities must be spelled out.
In this paper, we focus on this modeling for the open-vocabulary A2W end-to-end ASR system.


\section{Model description}\label{sec:model}
This section describes the baseline attention-based acoustic-to-word (A2W) model, the multi-task learning (MTL) framework with the acoustic-to-character (A2C) model, the recurrent neural network language model (RNNLM) integration to the attention-based models, and recovering OOV words with the A2C model.
Let $\bm{x}=(x_{1},\dots,x_{T})$ be the input speech frames of length $T$, $\bm{y}^{w}=(y^{w}_{1},\dots,y^{w}_{N})$ be the corresponding word-level transcription of length $N$, and $\bm{y}^{c}=(y^{c}_{1},\dots,y^{c}_{M})$ be the corresponding character-level transcription of length $M$.

\subsection{Baseline attention-based A2W model}\label{subsec:a2w}
In this paper, A2W models are built based on the attention-based encoder-decoder model \cite{attention_nips2015, las, attention_icassp2016}, which is an end-to-end sequence labeling model and can learn soft alignments between a variable-length input and a target sequence.
Attention-based models incorporate contextual information from the target label sequence in the decoder part unlike the CTC models, which is another end-to-end sequence labeling model and learn frame-level contexts in the encoder part.
We focus on the attention-based models as a baseline A2W model in this paper.
The A2W model is composed of three components: encoder, word-level decoder, and attention layer.

The encoder network consists of the stacked multiple layers of bidirectional long-short term memory (BLSTM) \cite{lstm1997} and transforms $\bm{x}$ into a distributed representation $\bm{h}=(h_1,\dots,h_T)$.

The decoder network consists of a single LSTM layer and generates the probability distribution of the lexical entries conditioned over all the previous outputs.
The decoder's hidden state $\bm{s}_{n}$ at each output timestep $n$ is updated as a function of the context vector $\bm{c}_{n}$, and previously output word $y^{w}_{n-1}$ (which is passed through the word embedding layer) recursively as follows:
\begin{align}
& y^{w}_{n} \sim Generate(\bm{s}_{n-1}, \bm{c}_{n}) \notag \\
& \bm{s}_{n} = Reccurency(\bm{s}_{n-1}, \bm{c}_{n}, y^{w}_{n})
\end{align}

The attention layer computes an attention distribution $\bm{\alpha}_{n}^{w}=(\alpha_{n,1}^{w},\dots,\alpha_{n,T}^{w})$, which is a relevance score over the entire encoder's outputs $\bm{h}$, and computes the context vector $\bm{c}_{n}$ by summing over the encoder's outputs $\bm{h}$ as follows:
\begin{align*}
& e_{n,t} = \bm{v}^{T}tanh(\bm{W}\bm{s}_{n-1} + \bm{V}\bm{h}_{t} + \bm{U}\bm{f}_{n,t} + \bm{b}) \\
& \bm{f}_{n} = \bm{F} * \bm{\alpha}^{w}_{n-1} \\
& \bm{\alpha}_{n}^{w} = softmax(\bm{e}_{n}) \\
& \bm{c}_{n} = \sum_{t=1}^{T}{\alpha_{n,t}^{w}\bm{h}_{t}}
\end{align*}
where $\bm{F}$, $\bm{W}$, $\bm{V}$, $\bm{U}$, $\bm{v}$, and $\bm{b}$ are trainable parameters and $*$ denotes convolutional operation, and $f_{n}$ is a convolutional feature from the previous attention distributions $\bm{\alpha}_{n-1}^{w}$.

The loss function is designed as the negative log-likelihood and used for parameter estimation:
\begin{align*}
L_{w}(\bm{x}, \bm{y}^{w}) = - \ln{P(\bm{y}^{w}|\bm{x})}
\end{align*}

\subsection{Multi-task learning with attention-based A2C model}\label{subsec:a2c}
To alleviate data sparseness issues of A2W models, the MTL with the A2C model is performed by sharing encoders' parameters.
As with the previous section, the A2C model is also built based on the attention-based model and has different parameters concerning the decoder and attention layer.
Although there is another choice of the CTC-based model for the A2C model as in \cite{ueno_icassp2018}, we adopt the attention-based model because character-level CTC models are more likely to misspell than attention-based models \cite{seq2seq_comparison_is17, baidu_seq2seq_asru2017}.
The character-level decoder can be connected to the arbitrary intermediate layer \cite{toshniwal2017multitask, analyzing_hidden_repr}.
The overall loss function is the linear interpolation of the negative log-likelihood between the A2W and A2C models by a tunable parameter $\lambda \ (0 \leq \lambda \leq 1)$:
\begin{align*}
L(\bm{x}, \bm{y}^{w}, \bm{y}^{c}) = \lambda L_{w}(\bm{x}, \bm{y}^{w}) + (1 - \lambda) L_{c}(\bm{x}, \bm{y}^{c})
\end{align*}
where $L_{c}$ is the negative log-likelihood of the A2C model.

\subsection{RNNLM integration}\label{subsec:shallow_fusion}
Although attention-based models explicitly model linguistic contexts in the decoder part, external LMs trained with a larger text corpus can provide a reliable probability distribution to the decoder.
The left-to-right beam search decoding with an external language model, which is referred to as \textit{shallow fusion} \cite{kannan2017analysis}, is performed to find the most probable word sequence $\bm{y}^{w*}$ based on the following criterion:
\begin{align*}
\bm{y^{w*}} = \argmax_{\bm{y}^{w}} \{ {\rm log} \ P_{a2w}(\bm{y}^{w}|\bm{x}) + \beta_{w} {\rm log} \ P_{wlm}(\bm{y}^w) \\
& \hspace{-27mm} + \gamma_{w} {\rm coverage} \}
\end{align*}
where $\beta_{w}$ and $\gamma_{w}$ are tunable parameters.
To use a narrow beam width and keep the decoding efficiency of the A2W model, scores by the external LM is added in the loop of the decoder network, not in the rescoring step.
The coverage terms are added to prevent long sequences composed of repeated tokens and calculated as follows:
\begin{align*}
{\rm coverage} = \sum_{t=1,...,T}[{\sum_{n=1,...,N}{\alpha^{w}_{n,t} > \tau}}]
\end{align*}
where $\tau$ is a threshold to receive a cumulative attention lager than its value.
We set $\tau$ to 0, and this also purges short hypotheses from candidates in the beam.

\begin{table*}[t]
  \centering
  \begingroup
  \caption{Recognition performances on Switchboard corpus (300h). SWB and CH represent Switchboard and CallHome subsets, respectively. \#OOV represents the number of detected OOV words. Beam search decoding was performed with $beam\_size=5$ in all experiments. The vocabulary size was about 11k. The OOV rates of SWB and CH test sets were 1.81 and 3.00, respectively.}
  \label{table:result_swbd}
  \scalefont{1.0}
    \vspace{-1mm}
    \begin{tabular}{|c|c|c|c|c|c|} \hline
     \multirow{2}{*}{Model} & \multirow{2}{*}{Resolving OOV} & \multirow{2}{*}{RNNLM} & SWB & CH & \multirow{2}{*}{\shortstack{Ave.\\WER}} \\ \cline{4-5}
      & & & WER (\#OOV) & WER (\#OOV) & \\ \hline \hline
      Word CTC & - & $\times$ & 20.26 (240) & 42.32 (358) & 31.29 \\ \hline


      \multirow{3}{*}{A2W} & - & $\times$ & 18.99 (154) & 38.46 (222) & 28.73 \\
      & - & 300h & 18.45 (319) & 38.49 (463) & 28.47 \\
      (baseline) & - & 2000h & 18.35 (322) & 38.13 (490) & 28.24 \\ \hline
      \multirow{6}{*}{A2W+A2C} & $\times$ & $\times$ & 18.35 (183) & 37.54 (267) & 27.95 \\
      & \checkmark & $\times$ & 18.18 (183) & 37.40 (267) & 27.79 \\
      & $\times$ & 300h & 17.76 (349) & 37.26 (513) & 27.51 \\
      & \checkmark & 300h & 17.43 (349) & 36.99 (513) & 27.21 \\
      & $\times$ & 2000h & 17.40 (346) & 37.00 (546) & 27.20 \\
      & \checkmark & 2000h & \bf 17.11 \rm (346) & \bf 36.71 \rm (546) & \bf 26.91 \\ \hline
    \end{tabular}
    \vspace{-1mm}
  \endgroup
\end{table*}

\subsection{Resolving OOV words}\label{subsec:resolving_oov}
In the MTL framework, the A2C model not only works as regularization effects but also provides additional information to the A2W model.
As with \cite{a2w_without_oov_asru2017, ueno_icassp2018}, in the inference stage, we refer to the A2C model's hypothesis $\hat{\bm{y}^{c}}=(\hat{y_{1}^{c}}, \cdots, \hat{y_{M}^{c}})$ when outputting OOV words and replace them with the corresponding space-separated word including $\hat{y_{m}^{c}}$ by computing the index $m$ where attention weights between the A2W and A2C models are most overlapped.
\begin{align*}
\{\overline{\bm{\alpha}_{m, i}^{c}}\}_{i} = \frac{\bm{\alpha}_{m, 2i}^{c} + \bm{\alpha}_{m, 2i+1}^{c}}{2} \\
m = \argmax_{m=1,\cdots,M}(\bm{\alpha}_{n}^{w} \cdot \overline{\bm{\alpha}_{m}^{c}})
\end{align*}
Since time resolutions of activations of the encoder attached to the A2W and A2C models are different due to the subsampling layers \cite{las}, frame-wise attention weight of A2C models $\bm{\alpha}^{c}$ is averaged between the adjacent two frames before multiplication.
Note that character-level hypotheses are obtained by greedy decoding to keep the decoding speed of the A2W models.

\section{Experimental evaluation}\label{sec:experiment}
\subsection{Switchboard corpus (300h)}\label{subsec:swbd_corpus}
\subsubsection{System settings}
We used the Switchboard corpus (LDC97S62) \cite{switchboard}, which contains about 300-hour conversational English telephone speech, as the training set.
Following data preparation in Kaldi recipe \cite{kaldi}, we reserved the first 4k utterances as a validation set separately.
Besides, we removed duplicated utterances ("yeah," "uh-huh" etc.) beyond a count threshold of 300.
The final training set has about 192k utterances.
For evaluation, we report results on Hub5 Eval2000 test set (LDC2002S09), which consists of two subsets, Switchboard (SWB) and CallHome (CH).

For A2W models, we restricted the vocabulary to words with at least five occurrences in the training set and replaced the rest to a single out-of-vocabulary (OOV) class.
The resulting vocabulary size was roughly 11k words, and the OOV rates of SWB and CH test sets were 1.81 and 3.00, respectively.
For A2C models, we used 47 character sets (26 alphabets, digits, hyphen, space, and end-of-sentence, etc.).
The input features were static 80-channel log-mel filterbank outputs computed with a 25ms window and shifted every 10 ms.
The features were normalized by the mean and the standard deviation on the speaker basis.
None of speaker adaption techniques was used.
The encoder consists of 5 stacked BLSTM layers with 320 memory cells per direction, and both word and character-level decoders consists of a single LSTM layer with 320 memory cells.
Subsampling was performed in \{1,2,4\}-th layers of the encoder to approximately equate sequence lengths to the number of the corresponding tokens in transcriptions.
The character-level decoder was attached to the 4-th layers of the encoder.
This resulted in 4 and 8-fold reduction of the encoder activations in the A2C and A2W models, respectively \cite{las, joint_ctc_attention, toward_better_decoding}.
Output words and characters were embedded to the fixed dimension of size 128 and 32, respectively.
The dropout ratio 0.2 was applied to the encoder, decoders, and embedding layers.
Training was performed on mini-batches of 50 or 60 utterances using Adam \cite{adam} with a learning rate of $1.0 \times 10^{-3}$ followed by SGD \cite{google_nmt} with a single GPU.
For fast and stable training, all utterances in the training set were sorted in the ascending order by their lengths in all training stage \cite{eesen, deepspeech2, direct_a2w_icassp2018}.
All weights were initialized with random values drawn from a uniform distribution with a range $[-0.1,0.1]$.
We also clipped the norms of gradients so that they had maximum absolute values of 5 \cite{gradient_clipping}.
The probabilities of scheduled sampling \cite{scheduled_sampling} and label smoothing \cite{rethinking_inception, toward_better_decoding} were 0.2 and 0.1, respectively.
We empirically set $\lambda=0.5$, $\beta_{w}=0.2$, $\gamma_{w}=0.4$ (w/o LM), and $\gamma_{w}=0.6$ (w/ LM), respectively.

RNNLMs were composed of two layers of unidirectional LSTM with 512 memory cells and have residual connections between two LSTM layers \cite{kurata2017empirical}.
Input and output embeddings were tied as in \cite{inan2016tying, press2016using}.
RNNLMs were optimized using back-propagation through time (BPTT) with a sequence length of 100.
We used the same transcriptions as the A2W models (300-hour) and also those appended with Fisher corpus (totally 2000-hour) for training RNNLMs.
The beam width was set to 5 in all the experiments.
All networks were implemented with a Pytorch framework \cite{pytorch}.

\begin{table*}[t]
  \centering
  \begingroup
  \caption{Recognition performances on CSJ (240h). \textit{eval3} is the out-of-domain test set. \#OOV represents the number of detected OOV words. Beam search decoding was performed with $beam\_size=5$ in all experiments. The vocabulary size was about 12.5k. The OOV rates of \textit{eval1}, \textit{eval2}, and \textit{eval3} were 1.24, 1.66, and 4.09, respectively.}
  \label{table:result_csj}
  \scalefont{0.95}
    \vspace{-1mm}
    \begin{tabular}{|c|c|c|c|c|c|c|} \hline
      \multirow{3}{*}{Model} & \multirow{3}{*}{Resolving OOV} & \multirow{3}{*}{RNNLM} & \multicolumn{2}{c|}{In-domain} & Out-of-domain & \multirow{3}{*}{\shortstack{Ave.\\WER}} \\ \cline{4-6}
      & & & \textit{eval1} & \textit{eval2} & \textit{eval3} & \\ \cline{4-6}
      & & & WER (\#OOV) & WER (\#OOV) & WER (\#OOV) & \\ \hline \hline
      Word CTC & - & $\times$ & 12.79 (352) & 11.12 (469) & 20.28 (662) & 14.73 \\ \hline
      A2C & - & $\times$ & 12.82 & 9.98 & 18.98 & 13.93 \\ \hline
      \multirow{3}{*}{A2W} & - & $\times$ & 12.89 (265) & 10.25 (299) & 19.70 (498) & 14.28 \\
      & - & 240h & 12.20 (437) & 9.73 (531) & 19.49 (761) & 13.80 \\
      (baseline) & - & 600h & 12.11 (443) & 9.65 (516) & 18.71 (759) & 13.49 \\ \hline

      \multirow{6}{*}{A2W+A2C} & $\times$ & $\times$ & 12.27 (252) & 9.96 (334) & 18.70 (521) & 13.64 \\
      & \checkmark & $\times$ & 12.06 (252) & 9.67 (334) & 17.99 (521) & 13.24 \\
      & $\times$ & 240h & 11.71 (441) & 9.40 (534) & 18.21 (782) & 13.11 \\
      & \checkmark & 240h & 11.27 (441) & 8.85 (534) & 17.20 (782) & 12.44 \\
      & $\times$ & 600h & 11.70 (429) & 9.29 (518) & 17.54 (788) & 12.85 \\
      & \checkmark & 600h & \bf 11.27 \rm (429) & \bf 8.77 \rm (518) & \bf 16.57 \rm (788) & \bf 12.21 \\ \hline
    \end{tabular}
    \vspace{-1mm}
  \endgroup
\end{table*}

\subsubsection{Results}\label{subsubsec:result_swbd}
The results are shown in Table \ref{table:result_swbd}.
The attention-based A2W model outperformed the word CTC model\footnote{As in \cite{direct_a2w_is17, direct_a2w_icassp2018}, we also confirmed that the initialization of the BLSTM encoder with phone CTC improved the performances of both word CTC and attention-based A2W model. The resulting WERs w/o LMs were 19.78/39.97 and 18.86/37.81 (SWB/CH) with $beam\_width=5$, respectively. However, the A2W model still outperformed the word CTC model.}\footnote{We did not use any speaker adaptation techniques such as i-vector based adaptation as in \cite{direct_a2w_is17, direct_a2w_icassp2018}. We assume that this is the major cause of the performance gaps between our results and those in \cite{direct_a2w_is17, direct_a2w_icassp2018} in CallHome subset.}.
This is because the decoder part in the attention-based A2W model captured richer linguistic contexts as mentioned in Section \ref{subsec:a2w}.
Both external RNNLMs trained with the original training data (300h) and that concatenated with the additional training data (2000h) improved the performances in Switchboard subset, and the latter improved the performances in CallHome subset.
Moreover, the number of detected OOV words in the hypotheses was increased by \textit{shallow fusion} with the external RNNLMs.
This is because the vocabulary of RNNLMs was limited to that of the A2W models, and high probabilities were assigned to the OOV class.
The MTL with the A2C model improved the baseline performance, and resolving OOV words boosted it as in our previous work \cite{ueno_icassp2018}.
External RNNLMs improved the performances of all models except for the one trained on{} 300h text data for the baseline A2W model.
The MTL approach encouraged the effectiveness of \textit{shallow fusion} thanks to alleviating data sparseness issues.
In addition, \textit{shallow fusion} enhanced the improvements by the OOV resolution.
This suggests that the external LM helps detect OOV words more accurately.
In summary, compared to the baseline A2W model with \textit{shallow fusion}, the further combination with the A2C model and the OOV resolution method yielded absolute 1.24 (6.75\% relative), and 1.42 (3.72\% relative) gains in Switchboard (SWB) and CallHome (CH) subsets, respectively.


\subsection{Corpus of spontaneous Japanese (CSJ)}\label{subsec:csj_corpus}
\subsubsection{System settings}
The Corpus of Spontaneous Japanese (CSJ) \cite{csj} is one of the largest Japanese spontaneous speech corpora.
The CSJ consists of about 600-hour spontaneous speech including academic and simulated lectures.
In this paper, we focus on the academic lectures which have been the primary target of ASR research using this corpus, consisting of about 240 hours of training data in total.
There are three evaluation sets (\textit{eval1}, \textit{eval2}, and \textit{eval3}), each of which is composed of 10 lectures and the \textit{eval3} set is regarded as an out-of-domain test set.
We picked up the first 4k utterances from the training set as a validation set separately following Kaldi recipe \cite{kaldi}.
The final training set has about 155k utterances.

For the A2W models, we restricted the vocabulary to words which occurred at least five times in the training set and replaced the rest to a single OOV class.
The resulting vocabulary size was about 12.5k words, and the OOV rates of \textit{eval1}, \textit{eval2}, and \textit{eval3} were 1.24, 1.66, and 4.09, respectively.
For the A2C model, we used the 2820 kinds of standard Japanese characters including kanji, hiragana, and katakana characters, alphabets, digits, noise, space, and the end of sentence mark.
Output words and characters were embedded in the fixed dimension of size 128 and 64, respectively.
The topology and training scheme of RNNLMs were the same as in Section \ref{subsec:swbd_corpus}.
The rest of the configurations was the same as in Section \ref{subsec:swbd_corpus}.

RNNLMs were composed of two layers of unidirectional LSTM with 512 memory cells and have residual connections between two LSTM layers \cite{kurata2017empirical}.
Input and output embeddings were tied as in \cite{inan2016tying, press2016using}.
RNNLMs were optimized using back-propagation through time (BPTT) with a sequence length of 100.
We used the same transcriptions as ASR models (240-hour) and also those appended with additional text data of the simulated lectures (totally 600-hour) for training RNNLMs.

\subsubsection{Results}
The results are shown in Table \ref{table:result_csj}.
As in Section \ref{subsubsec:result_swbd}, the attention-based A2W model outperformed the word CTC model.
In contrast, the A2C model outperformed the A2W model.
This is because the length of characters per word in Japanese is shorter than that in English, so it is easy for the A2C model to capture linguistic dependencies from the target character sequence.
The MTL of the A2W model with the A2C model improved the performance, and outperformed the A2C model.
Resolving OOV words further improved the performance, which is consistent with results in Section \ref{subsubsec:result_swbd}.

Both external RNNLMs trained with the original training data (240h) and that concatenated with the additional training data (600h) improved performances in all test sets in proportion to the training data size for LMs.
Adding the out-of-domain data to the training data for the external RNNLM alleviated the domain mismatch in \textit{eval3} test set to some extent.
The MTL alleviated data sparseness problems, and then the effect of \textit{shallow fusion} was emphasized.
\textit{shallow fusion} also increased the number of detected OOV words in the hypotheses in this corpus, and this left room for the improvements by recovering OOV words by the A2C model.
In summary, the combination of the MTL with the A2C model, resolving OOV words and \textit{shallow fusion} with the external RNNLM showed the best WER in three test sets, especially for the out-of-domain scenario (\textit{eval3}).
Compared to the baseline A2W model with \textit{shallow fusion}, our best model yielded absolute 0.84 (6.93\% relative), 0.88 (9.11\% relative) and 2.14 (11.43\% relative) gains in each test subset, respectively.

\begin{figure}[t]
    \centering
    \vspace{-1mm}
    \includegraphics[width=1.0\hsize,scale=1.0]{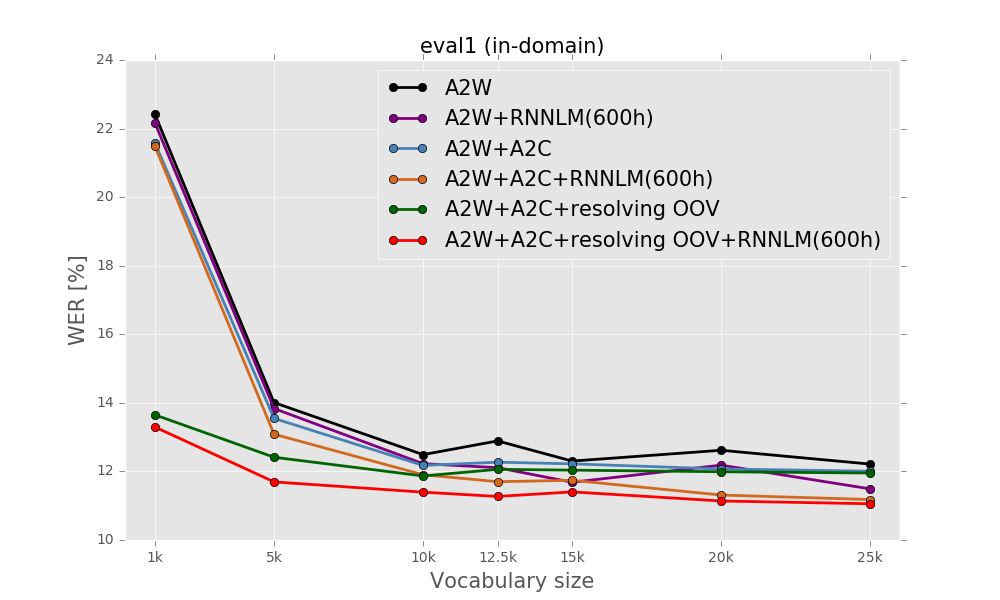}
    \caption{Recognition performances with various vocabulary sizes in \textit{eval1} test set. \textit{eval1} is regarded as an in-domain test set.}
    \label{fig:vocab_eval1}
    \vspace{-1mm}
\end{figure}

\begin{figure}[t]
    \centering
    \vspace{-1mm}
    \includegraphics[width=1.0\hsize,scale=1.0]{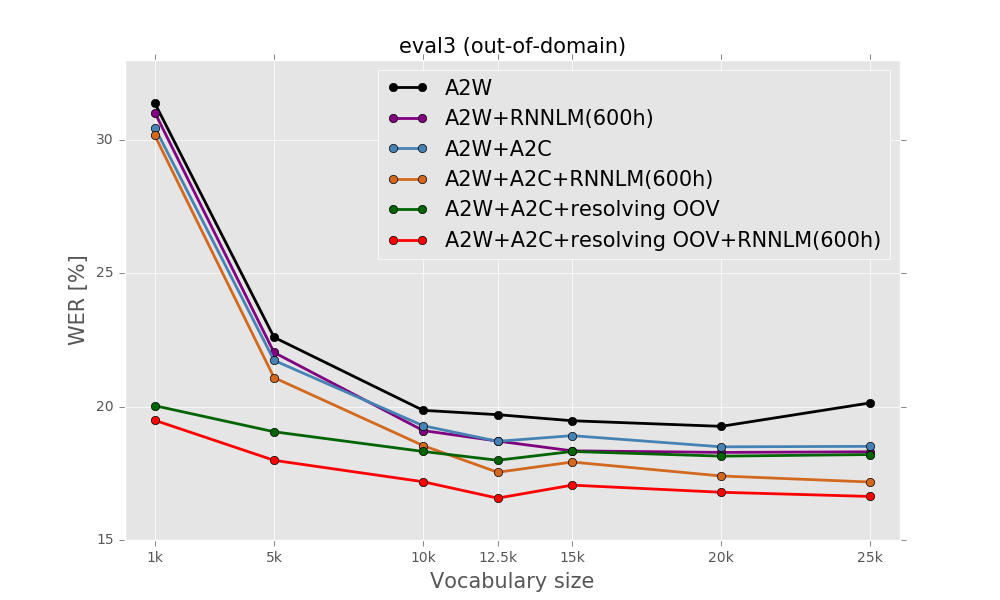}
    \caption{Recognition performances with various vocabulary sizes in \textit{eval3} test set. \textit{eval3} is regarded as an out-of-domain test set.}
    \label{fig:vocab_eval3}
    \vspace{-1mm}
\end{figure}

Next, we changed the vocabulary size of the A2W models (see Figure \ref{fig:vocab_eval1} and \ref{fig:vocab_eval3}).
The OOV rates in each vocabulary are shown in Table \ref{table:oov_csj_vocab}.
With the smaller vocabulary size, WER was drastically degraded due to the increase of the OOV rates in the test sets.
However, the MTL approach with OOV resolution mitigated this problem and was robust to the vocabulary size.
In the A2W models, the gain by the external RNNLMs was trivial with the small vocabulary.
In contrast, external RNNLMs were always effective in case of using the OOV resolution even with the very small vocabulary such as 1k and 5k.
The best results were obtained with vocabulary 15k, but the gaps of the performances between 5k and 15k were 0.30 and 0.94 in \textit{eval1} and \textit{eval3} test sets, respectively.
Therefore, we can reduce the vocabulary size three times with the small performance degradation.

\begin{figure}[t]
    \centering
    \vspace{-1mm}
    \includegraphics[width=1.0\hsize,scale=1.0]{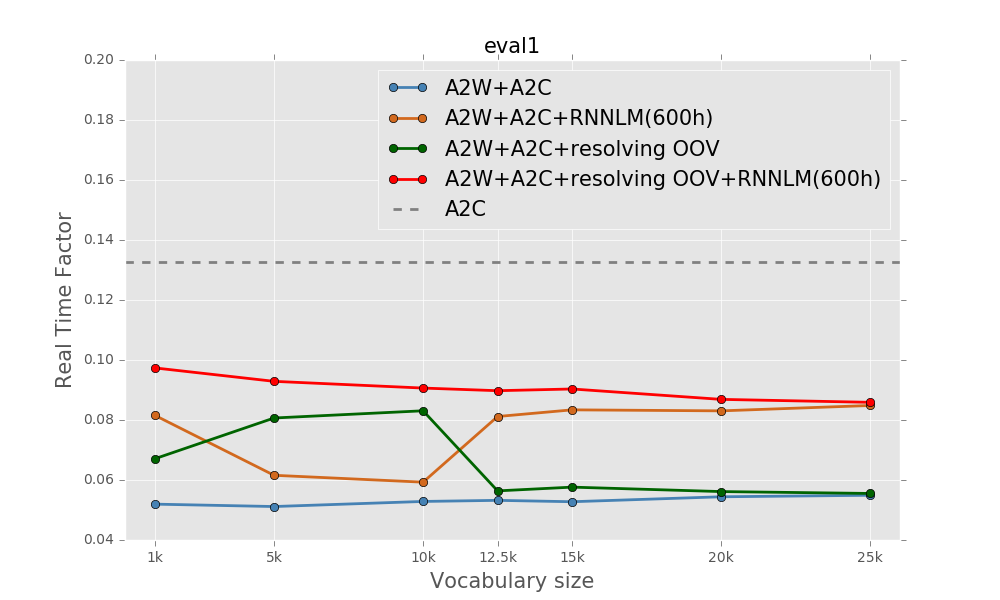}
    \caption{Real time factor in \textit{eval1} test set.}
    \label{fig:rtf_eval1}
    \vspace{-1mm}
\end{figure}

Finally, the real time factor (RTF) of the A2W models in \textit{eval1} test set are shown in Figure \ref{fig:rtf_eval1}.
Decoding was performed with a single NVIDIA Titan GPU.
Our attention-based models use the bidirectional encoders, but RTF is small enough for the real-time usage.
When using a large vocabulary, there is few additional time for resolving OOV words with the external RNNLM because there are few OOV words.
In contrast, when using a small vocabulary, the costs of the OOV resolution is more expensive than those of using external RNNLMs.
However, note that all of the A2W models are always faster than the A2C model although some of them use the external RNNLM during decoding.


\begin{table}[t]
  \centering
  \begingroup
  \caption{The OOV rates of in-domain (\textit{eval1}) and out-of-domain (\textit{eval3}) test sets in CSJ corpus (\%).}
  \label{table:oov_csj_vocab}
  \scalefont{1.0}
    \vspace{-1mm}
    \begin{tabular}{|c|c|c|c|} \hline
      Vocabulary size & \textit{eval1} & \textit{eval3} \\ \hline \hline
      1k & 11.87 & 18.86 \\ \hline
      5k & 3.14 & 8.20 \\ \hline
      10k & 1.65 & 4.79 \\ \hline
      15k & 1.30 & 3.64 \\ \hline
      20k & 0.99 & 3.10 \\ \hline
      25k & 0.92 & 2.69 \\ \hline
    \end{tabular}
    \vspace{-1mm}
  \endgroup
\end{table}

\section{Conclusions}\label{sec:conclusions}
We have addressed an issue that the acoustic-to-word (A2W) model tends to incorrectly recognize OOV words as in-vocabulary words.
Joint decoding with the external language model helps the A2W model detect OOV words more accurately because it has more reliable linguistic information.
These OOV words can be recovered by the character-level decoder which attached to the same encoder as the A2W model in the multi-task learning (MTL) framework.
We experimentally confirmed that external LMs encouraged the OOV prediction, and recovering OOV words further improved the performance.
We also found that MTL alleviates data sparseness issues to some extent, and then the effectiveness of the LM integration is enhanced.
In addition, by resorting the recognition of rare words to the character-level decoder, the A2W models can work with a small vocabulary size.



\bibliographystyle{IEEEbib}
\bibliography{slt2018_reference}

\end{document}